\documentclass{article}

\usepackage{microtype}
\usepackage{graphicx}
\usepackage{xcolor}
\usepackage{empheq}
\usepackage{subfigure}
\usepackage{booktabs} 
\usepackage{nicefrac}

\usepackage{amsmath}
\usepackage{amssymb}
\usepackage{amsthm}
\usepackage{gensymb}
\usepackage{booktabs,arydshln}
\usepackage[frozencache,cachedir=.]{minted}
\usepackage{fontawesome}
\usepackage{reledmac}
\arrangementX[A]{paragraph}
\let\footnote\footnoteA
\makeatletter
\def\adl@drawiv#1#2#3{%
        \hskip.5\tabcolsep
        \xleaders#3{#2.5\@tempdimb #1{1}#2.5\@tempdimb}%
                #2\z@ plus1fil minus1fil\relax
        \hskip.5\tabcolsep}
\newcommand{\cdashlinelr}[1]{%
  \noalign{\vskip\aboverulesep
           \global\let\@dashdrawstore\adl@draw
           \global\let\adl@draw\adl@drawiv}
  \cdashline{#1}
  \noalign{\global\let\adl@draw\@dashdrawstore
           \vskip\belowrulesep}}
\makeatother
\usepackage{stmaryrd}

\newcommand*{\sunderbin}[3][0pt]{%
  \mathbin{\renewcommand*{\arraystretch}{0}%
    \begin{array}[t]{@{}c@{}}%
      #3\\[{#1}]%
      \mathclap{\scriptstyle #2}%
    \end{array}}%
}

\newtheorem{theorem}{Proposition}

\usepackage{hyperref}

\usepackage[accepted]{icml2021}

\icmltitlerunning{The Earth Mover's Pinball Loss}

\begin{document}

\twocolumn[
\icmltitle{The Earth Mover's Pinball Loss: \\ Quantiles for Histogram-Valued Regression}




\begin{icmlauthorlist}
\icmlauthor{Florian List}{sy}
\end{icmlauthorlist}

\icmlaffiliation{sy}{The University of Sydney, Sydney Institute for Astronomy, School of Physics, A28, NSW 2006, Australia}
\icmlcorrespondingauthor{Florian List}{florian.list@sydney.edu.au}


\icmlkeywords{histogram, uncertainty quantification, quantile regression, loss function, optimal transport,  pinball loss, Earth Mover's distance, Wasserstein distance} 

\vskip 0.3in
]



\printAffiliationsAndNotice{}  

\begin{abstract}
Although ubiquitous in the sciences, histogram data have not received much attention by the Deep Learning community. Whilst regression and classification tasks for scalar and vector data are routinely solved by neural networks, a principled approach for estimating histogram labels as a function of an input vector or image is lacking in the literature.
We present a dedicated method for Deep Learning-based histogram regression, which incorporates cross-bin information and yields \emph{distributions} over possible histograms, expressed by $\tau$-quantiles of the cumulative histogram in each bin. The crux of our approach is a new loss function obtained by applying the pinball loss to the cumulative histogram, which for 1D histograms reduces to the Earth Mover's distance (EMD) in the special case of the median ($\tau = 0.5$), and generalizes it to arbitrary quantiles. We validate our method with an illustrative toy example, a football-related task, and an astrophysical computer vision problem. We show that with our loss function, the accuracy of the predicted \emph{median} histograms is very similar to the standard EMD case (and higher than for per-bin loss functions such as cross-entropy), while the predictions become much more informative at almost no additional computational cost. \href{https://github.com/FloList/EMPL}{\faGithub}
\end{abstract}

\section{Introduction}
\label{sec:introduction}

\begin{figure}[htb]
    \centering
    \resizebox{1\columnwidth}{!}{
    \includegraphics{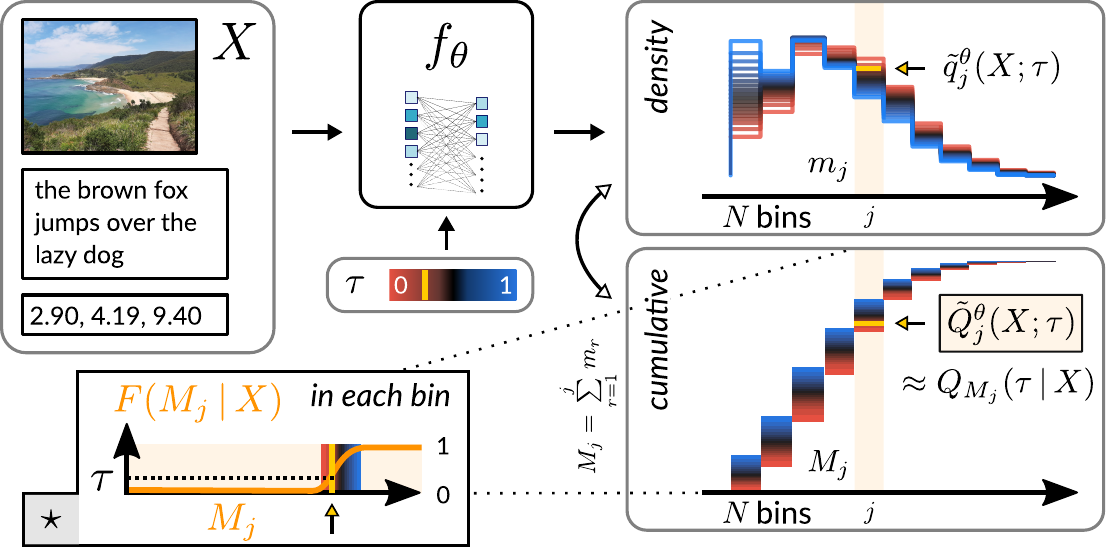}
    }
    \caption{Comic of our method for Deep Learning-based histogram regression: a neural network (NN) $f_\theta$ with weights $\theta$ is trained to estimate arbitrary $\tau$-quantiles of the \emph{cumulative} histogram $(M_j)_{j=1}^N$ in each bin $j \in \{1, \ldots N\}$ (treated as $N$ random variables), conditional on an input $X$. That is, the $j$-th NN output $\tilde{Q}_j^\theta(X; \tau) \approx Q_{M_j}(\tau \, | \, X)$ is an approximation of the true quantile function of $M_j$ given $X$ (i.e., the pre-image of $\tau$ under the CDF $F(M_j \, | \, X)$, see panel $\star$). Thus, the entire \emph{distribution} of possible histograms is learned as a function of $X$. The values of the associated \emph{density} histogram $(\tilde{q}_j)_{j=1}^N$, defined as $\tilde{q}_j = \tilde{Q}_j - \tilde{Q}_{j-1}$, increase early (late) for high (low) quantile levels $\tau$.}
    \label{fig:sketch}
\end{figure}

Histograms, i.e. approximate representations of the distribution of numerical data obtained by binning the data into adjacent, non-overlapping bins, are frequently used across the disciplines. Typical examples are precipitation histograms in meteorology (e.g. \citealt{Nicholls1997}), population pyramids in demographics and ecology (e.g.  \citealt{weeks2020population}, and color histograms in photography and image processing (e.g. \citealt{223129}). While the invention of histograms is commonly attributed to Karl Pearson \citep{Pearson1895}, the use of bars for the representation of data can be traced back to the Middle Ages \cite{Oresme}. In Deep Learning, histograms appear in different contexts: they can be used as neural network (NN) \emph{inputs} (e.g. \citealt{Saadl2009, Rebetez2016}), different variants of \emph{hidden} histogram layers have been proposed \citep{10.1007/978-3-319-46448-0_15, Sedighi2017, Peeples2020}, block-wise histograms have been employed for feature pooling in \citet{Chan2015}, and histogram loss functions were introduced in \citet{Ustinova2016, Zholus2020}. Furthermore, histograms of the trainable NN weights in different layers can shed light on whether the NN training is progressing properly.
\par In contrast, the task of regressing histogram \emph{labels} based on an input vector (or image) $X$ using NNs has not received much attention to date. This is despite the great potential of Deep Learning for identifying complex and nonlinear relations between an input $X$ and an associated histogram, which is a common problem in many areas. For instance, \citet{Bellerby2007} used a NN to predict rainfall histograms based on satellite-derived input data, \citet{Liu2020a} considered the Deep Learning-based estimation of dose-volume histograms for radiotherapy planning, and \citet{Sharma2020} presented a CNN for the recovery of object size histograms from images, taking fly larvae and breast cancer cell data as examples.
\par Clearly, an ad-hoc approach is to treat each bin separately and to use a standard \emph{regression} loss function such as the $l^1$ or $l^2$ error (mean absolute error and mean squared error, respectively), or a standard \emph{classification} loss (softmax activation + cross-entropy loss); the latter assuming that the histograms sum up to unity (with the true label vector components lying anywhere in $[0, 1]$ instead of $\{0, 1\}$ as in the case of one-hot coded class labels for an actual classification problem).
However, a drawback of these approaches is that the inherent ordering of the histogram bins is ignored, and cross-bin correlations are thus disregarded. Such an ordering may also be present in classification tasks: namely, whenever a continuous variable is discretized into bins, such as for image-based age estimation with labels ``child'', ``adult'', and ``senior'', as opposed to viewing the problem as a regression task, where the age is estimated as a number (e.g. in years). For these scenarios, \citet{Hou2016} suggested the use of the (squared) Earth Mover's distance (EMD; \citealt{Rubner2000}) as a loss function. The EMD measures the minimal amount of work needed to transform a distribution into another, and therefore penalizes the NN more when placing probability mass into bins far from the correct one, whereas the cross-entropy loss considers each bin in isolation. Another cross-bin loss function with a similar motivation is the Cumulative Jenssen--Shannon divergence \citep{Nguyen2015, Jin2018}. The important difference between ordered classification and our setting is, however, that we are interested in the \emph{entire histogram}, not only in the argmax, which becomes the estimated class label for ordered classification, while the remaining estimated class probabilities are typically discarded. Therefore, we need the NN to correctly predict the value (or even the entire distribution of potential values) \emph{in every bin}. This bears similarity to concepts such as Label Distribution Learning (LDL; \citealt{Geng2016}), where the entire label distribution is relevant. However, LDL does not assume an underlying ordering of the labels, and categorical labels (e.g. ``sky'', ``plant'', ``mountain'') are supported, whereas our approach is specifically tailored to histograms.
\par In this work, we introduce a method for the NN-based estimation of conditional histograms. Since each input $X$ can potentially belong to an entire \emph{distribution} of output histograms, regressing a single ``mean histogram'' is often not sufficient, especially in applications where the \emph{range} of possible values in each bin for a specific input may have severe implications as in e.g. medicine. For this reason, we base our approach in \emph{quantile regression} \citep{Koenker1978}, enabling us to estimate arbitrary quantiles of the cumulative histogram in each bin. Specifically, we naturally extend the EMD for 1D distributions by allowing for asymmetry, in analogy to the pinball loss being an asymmetric generalization of the $l^1$ loss. In Sec.~\ref{sec:EM_pinball_loss}, we introduce the EMD and the pinball loss, and we define the \emph{Earth Mover's Pinball Loss} (EMPL) by combining the two. Then, we demonstrate the effectiveness of our method in three scenarios in Sec.~\ref{sec:results}: first, we consider a toy example that can be phrased in terms of drawing numbered balls from an urn. Second, we consider an application in sports and use the EMPL to estimate league table positions. Finally, we consider a problem from $\gamma$-ray astronomy: the recovery of the brightness distributions of point-sources from photon-count maps. We conclude this work in Sec.~\ref{sec:conclusions}.

\section{Earth Mover's Pinball Loss}
\label{sec:EM_pinball_loss}
In this section, we introduce the EMPL for the task of histogram-valued quantile regression. We start by formally defining the optimization problem to be solved. Then, we recall the definitions of the EMD and the pinball loss, and proceed by defining the EMPL as a natural asymmetric extension of the EMD, which allows us to obtain quantiles for cumulative distribution functions (and hence for cumulative histograms in the discrete setting).

\subsection{Problem formulation}
We consider the task of learning a mapping from an independent (random) variable $X$ to a corresponding distribution over output histograms with $N \in \mathbb{N}$ bins. We express this distribution over output histograms in terms of their quantiles, which has the advantage that no closed form for the distribution needs to be specified, making our method suitable for highly non-Gaussian and multimodal distributions. Recall that for a real-valued random variable $Y$ with a strictly monotonic CDF $F_Y(y) = P(Y \leq y)$ and $\tau \in (0, 1)$, the $\tau$-th quantile of $Y$ is defined as
\begin{equation}
    Q_Y(\tau) = F^{-1}_Y(\tau) = \inf \{y : F_{Y}(y) \geq \tau \}.
\end{equation}
\par For the sake of simplicity, we assume that the histogram values $(m_j)_{j=1}^N$ are normalized, i.e. $m_j \in [0, 1]$ and $\sum_{j = 1}^{N} m_j$ = 1, but our approach can easily be extended to arbitrary (non-negative) histograms by appending the total histogram count before normalization as an additional NN output. Further, we define the \emph{cumulative} histogram by setting $M_j = \sum_{r = 1}^j m_r$, and we write $M = (M_j)_{j=1}^N$.
\par Let $f_\theta$ be a NN with trainable weights $\theta$, whose task is to predict $\tau$-quantiles of the cumulative histogram $\tilde{Q}^\theta(X; \tau)$. Our goal is to determine optimal parameters $\theta^*$ such that
\begin{equation}
    \theta^* = \underset{\theta}{\operatorname{arg \, min}} \ 
    \sunderbin[2pt]{X\!, \tau}{\mathbb{E}} \
    \left[  
        \left\|Q_M(\tau \, | \, X) - \tilde{Q}^\theta(X; \tau) \right\|_1 
    \right],
\label{eq:problem_formulation}
\end{equation}
where $Q_M(\tau \, | \, X)$ is the vector-valued function that gathers the true $\tau$-quantiles of the cumulative histogram $M$ from all the bins $j = 1, \ldots, N$, given $X$. The expected value is taken over the input $X \sim P_X$ and uniformly over the quantile levels $\tau \sim U(0, 1)$, and $\| \cdot \|_1$ is the $l^1$-norm on $\mathbb{R}^N$ that sums up the approximation errors from all the bins to a single number. A sketch of the histogram regression process is shown in Fig.~\ref{fig:sketch}.

\subsection{Earth Mover's distance}
As a first step towards solving Eq.~\eqref{eq:problem_formulation}, we introduce the EMD \citep{Rubner2000}, which is a distance measure between probability distributions rooted in the Optimal Transportation problem \citep{Villani2009}. As will be seen later, minimizing the EMD between the true and estimated histograms yields NN weights $\theta$ such that $\tilde{Q}^{\theta}(X; \tau) \approx Q_M(\tau \, | \, X)$ for the specific case of the \emph{median} ($\tau = 0.5$).
\par Intuitively, the EMD measures the minimal amount of work that needs to be done in order to turn the ``pile of dirt'' (or earth) given by the PDF of distribution $u$ into that of another distribution $v$.
The definition of the EMD is in terms of ``signatures'', defined as sets of clusters each of which contains a certain amount of mass, and permits different total masses for different signatures. In the case of equal masses, however, it can be shown \citep{Levina2001} that the EMD is equivalent to the Wasserstein distance \citep{villani2003topics, Arjovsky2017}.
\par Interpreting the normalized histograms $(m_j)_{j=1}^N$ as discretizations of continuous PDFs, we directly introduce the EMD in the continuous framework of the Wasserstein distance, which formally reads as follows:
\par For $d \in \mathbb{N}$, $p \in [1, \infty)$, let $u, v$ be Borel probability measures on $\mathbb{R}^d$ with finite $p$-moments. Then, the $p$-Wasserstein distance is defined as \citep{Villani2009}
\begin{equation}
    W_p(u, v) = \left( \inf_{\pi \in \Gamma(u, v)} \int_{\mathbb{R}^d \times \mathbb{R}^d} \|x - y\|^p \, d\pi(x, y) \right)^{1/p},
    \label{eq:wasserstein}
\end{equation}
where $\Gamma(u, v)$ is the collection of joint probability measures on $\mathbb{R}^d \times \mathbb{R}^d$ with marginals $u$ and $v$ for the first and second argument, respectively. 
The definition of the $p$-Wasserstein distance does \emph{not} require the measures $u$ and $v$ to be absolutely continuous w.r.t. the Lebesgue measure $\lambda$ and is equally well-defined for discrete measures such as the Dirac measure, in which case the aforementioned notion of discrete clusters containing points can be recovered.
\par In this work, we restrict ourselves to the $1$-Wasserstein distance in the one-dimensional case, i.e. $p = 1$ and $d = 1$, in which the otherwise difficult calculation of the Wasserstein distance is greatly simplified and admits the following closed-form solution \citep{Ramdas2017}:
\begin{equation}
    W_1(u, v) = \int_\mathbb{R} |U(t) - V(t)| \ dt,
\label{eq:W1_with_CDFs}
\end{equation}
where $U$ and $V$ are the CDFs of $u$ and $v$, respectively, implying that the $1$-Wasserstein distance is simply given as the $L^1$-distance between the CDFs of the two distributions in the 1D case, which arises from a notion of monotonicity that the optimal transport plan needs to satisfy (see \citealt{Ramdas2017}). 

\subsection{Pinball loss}
Now, we turn towards the problem of quantile regression. For a scalar random variable $Y$, let $\tilde{y}$ be an approximation of the true quantile function $Q_Y(\tau)$. A suitable distance for comparing $\tilde{y}$ with observed values $y$ is the pinball loss function \citep{Fox1964, Koenker1978, Koenker2001, ferguson2014mathematical}, defined as
\begin{equation}
    \begin{aligned}
    \mathcal{L}_\tau^\text{pin}(y, \tilde{y}) &= (y - \tilde{y}) \left(\tau - \mathbb{I}_{(y < \tilde{y})}\right) \\ &=
    \begin{cases}
    \tau (y - \tilde{y}), & \text{if } y \geq \tilde{y}, \\
    (\tau - 1) (y - \tilde{y}), & \text{if } y < \tilde{y}.
    \end{cases}
\label{eq:pinball_loss}
    \end{aligned}
\end{equation}
The pinball loss is constructed in such a way that its expectation is minimized by $\tilde{y} = Q_Y(\tau)$, which follows immediately from setting the derivative of the expected loss function w.r.t. $\tilde{y}$
\begin{equation}
\begin{aligned}
    \frac{\partial \mathbb{E}_Y[\mathcal{L}_{\tau}^\text{pin}(Y, \tilde{y})]}{\partial \tilde{y}} &= (1 - \tau) F_Y(\tilde{y}) - \tau (1 - F_Y(\tilde{y})) \\
    &= F_Y(\tilde{y}) - \tau,
\label{eq:quantile_minimizes_pinball}
\end{aligned}
\end{equation}
to zero, which yields the minimum at $\tilde{y} = Q_Y(\tau)$.

\subsection{Quantile regression for histograms}
Having introduced the EMD between probability distributions and the pinball loss for quantile regression, we now combine the two for the task of \emph{histogram-valued regression}. We proceed in the continuous framework and subsequently consider the discrete case (i.e., histograms instead of PDFs).
\par For probability measures $u$ and $v$, we define the EMPL as
\begin{equation}
    \mathcal{L}_\tau(u, v) = \int_\mathbb{R} \left(U - V\right) \left(\tau - \mathbb{I}_{(U < V)}\right) \, dt,
    \label{eq:EM_pinball_loss}
\end{equation}
where $U$ and $V$ are again the CDFs of $u$ and $v$, respectively, and we suppress the argument $t$ for brevity. This loss function can be viewed as an asymmetric extension of Eq.~\eqref{eq:W1_with_CDFs} in the spirit of the pinball loss in Eq.~\eqref{eq:pinball_loss}, with the asymmetry governed by the quantile level of interest $\tau$. We can decompose the integral into two regions and write
\begin{equation}
    \mathcal{L}_\tau(u, v) = (1 - \tau) \int_{U < V} \hspace{-0.2cm} \left| U - V \right| \, dt + \tau \int_{U \geq V} \hspace{-0.2cm} \left| U - V \right| \, dt,
\end{equation}
from which the following bounds in terms of the 1-Wasserstein distance follow immediately:
\begin{equation}
    \eta_{-} W_1(u, v) \leq \mathcal{L}_\tau(u, v) \leq \eta_{+} W_1(u, v) \leq W_1(u, v),
\end{equation}
where $\eta_{-} = \min\{\tau, 1 - \tau\}$ and $\eta_{+} = \max\{\tau, 1 - \tau\}$. Note in particular that for the median ($\tau = 0.5$), one obtains
\begin{equation}
    \mathcal{L}_{0.5}(u, v) = \frac{1}{2} \int_\mathbb{R} |U - V| \, dt = \frac{1}{2} \, W_1(u, v),
\label{eq:EMPL_generalises_EMD}
\end{equation}
and the 1-Wasserstein distance in the case $d = 1$ is recovered up to the factor of $1/2$ (see Eq. \eqref{eq:W1_with_CDFs}). For $\tau \neq 0.5$, the EMPL is not symmetric and generally $\mathcal{L}_\tau(u, v) \neq \mathcal{L}_\tau(v, u)$, but rather $\mathcal{L}_\tau(u, v) = \mathcal{L}_{1 - \tau}(v, u)$. Figuratively speaking, one could think of moving probability mass up or down a hill whose slope is determined by the quantile level of interest, making it more difficult to move probability mass upwards than downwards.
\par In the discrete setting, Eq. \eqref{eq:EM_pinball_loss} becomes
\begin{equation}
    \mathcal{L}_\tau(u, v) = \frac{1}{N} \sum_{j=1}^{N} \left[ \left(U_j - V_j\right) \left(\tau - \mathbb{I}_{(U_j < V_j)}\right) \right],
    \label{eq:EM_pinball_loss_discrete}
\end{equation}
where $U_j = \sum_{r=1}^j u_r$ and similarly for $V_j$. We remark that the EMPL as defined in Eq.~\eqref{eq:EM_pinball_loss_discrete} implicitly assumes the ``distance'' $d_{ij}$ between two bins $i$ and $j$ in the notion of ``work'' when moving probability mass to be proportional to the distance between the bin indices, i.e. $d_{ij} \propto |i - j|$. For example, if one uses equally spaced (logarithmically spaced) bins in terms of the underlying variable $R$ whose distribution is described by the histogram, the distance scales linearly with $R$ (with $\log R$).
\par Coming back to the problem formulation, it now becomes apparent that training a NN using the EMPL as defined in Eq.~\eqref{eq:EM_pinball_loss_discrete} provides an (approximate) solution to Eq.~\eqref{eq:problem_formulation}: 

\begin{theorem}
For each fixed input $X = x$ and quantile level $\tau \in (0, 1)$, a NN returning the conditional quantiles of the cumulative histogram, i.e. $\tilde{Q}^\theta(x; \tau) = Q_{M}(\tau \, | \, x)$, minimizes the expected $\tau$-EMPL between observed cumulative histograms $\bar{M}(x)$ and the NN prediction $\tilde{Q}^\theta(x; \tau)$.
\vspace{-0.4cm}
\begin{proof}
This follows directly from plugging $U = \bar{M}(x)$ and $V = \tilde{Q}^\theta(x; \tau)$ into Eq.~\eqref{eq:EM_pinball_loss_discrete} and using the same argument as in Eq.~\eqref{eq:quantile_minimizes_pinball} for each bin $j \in \{1, \ldots N\}$.
\end{proof}
\end{theorem}
\vspace{-0.2cm}
An in-depth theoretical (convergence) analysis of the EMPL is beyond the scope of this paper and left to future work; however, our results in Sec.~\ref{sec:results} are encouraging and confirm its suitability for diverse practical use cases.

\subsection{Implementation details}
\label{subsec:implementation}
 The expectation over the inputs $X$ in Eq.~\eqref{eq:problem_formulation} is approximated as usual by training the NN on a large number of representative training samples $(X_s)_{s=1}^S$ with $X_s \sim P_X$ by means of a mini-batch gradient descent method. As for the expectation over the quantile level $\tau$, we follow \citet{Tagasovska2018} and draw an individual quantile level $\tau$ for each input $X_s$ from a uniform distribution $\tau \sim U(0, 1)$ during the NN training. Compared with NNs that are trained for a single quantile level $\tau$, the authors of that work showed that simultaneously estimating all the quantile levels greatly reduces \emph{quantile crossing} in the case of scalar-valued NNs. The quantile levels $\tau$ appear at two places in the NN: 1) they are fed as an additional NN input in order for the NN to know which quantile level shall be estimated, and 2) they are used in the computation of the loss.
\par In practice, we obtain $\tilde{Q}$ by 1) estimating $N$ logits $(\tilde{l}_j)_{j=1}^N$ (one per bin), 2) applying the softmax function $\tilde{q}_j = \operatorname{softmax}(\tilde{l})_j$, which yields a normalized \emph{density} histogram $(\tilde{q}_j)_{j=1}^N$, and 3) setting $\tilde{Q}_j = \sum_{r=1}^j \tilde{q}_r$, which enforces $\tilde{Q}_N = 1$. This implies that the NN prediction is properly normalized for all $\tau$; moreover, monotonicity of the predicted cumulative histograms for each fixed $\tau$ is guaranteed because of $\tilde{Q}_j = \tilde{Q}_{j-1} + \tilde{q}_j$ with $\tilde{q}_j \in (0, 1)$. The monotonicity of the quantiles \emph{within} each bin is not strictly enforced, but it is encouraged by Eq.~\eqref{eq:EM_pinball_loss_discrete}. Although we rarely ever encountered quantile crossing in our experiments once the NN is trained, crossing penalty terms as proposed by \citet{Takeuchi2006} could be incorporated in our framework without difficulty.
\par We also consider a Smoothed EMPL that is differentiable everywhere, derived by replacing the pinball loss by the smooth approximation proposed in \citet{Zheng2011} (and applied to NNs in \citealt{Hatalis2019}), which yields
\begin{equation}
    \begin{aligned}
    \mathcal{L}^\alpha_\tau(u, v) &= \frac{1}{N} \sum_{j=1}^N \! \Bigg{[}\tau \left(U_j - V_j\right) \\
    &\qquad + \alpha \, \log \left(1 + \exp\left(\frac{V_j - U_j}{\alpha}\right) \right) \Bigg{]},
    \end{aligned}
\end{equation}
where $\alpha > 0$ is a smoothing parameter. In the limit $\alpha \searrow 0$, $\mathcal{L}_\tau(u, v)$ is recovered, and for any $\alpha > 0$, $\mathcal{L}^\alpha_\tau(u, v) - \mathcal{L}_\tau(u, v)$ is $\tau$-independent, as follows immediately from the definitions. Note that $\log \left(1 + \exp(\cdot)\right)$ is the softplus function, which is readily available in most machine learning libraries.
An alternative approach that also provides differentiability everywhere is to consider an ``$L^2$-version'' of the EMPL,
leading to the estimation of $\tau$-expectiles rather than $\tau$-quantiles \citep{Aigner1976, Newey1987}, which generalize the \emph{mean} instead of the median and occasionally find use in financial risk estimation, but lack intuitive interpretability.

\section{Results}
\label{sec:results}
\begin{figure}[!htb]
    \centering
    \resizebox{1\columnwidth}{!}{
    \includegraphics{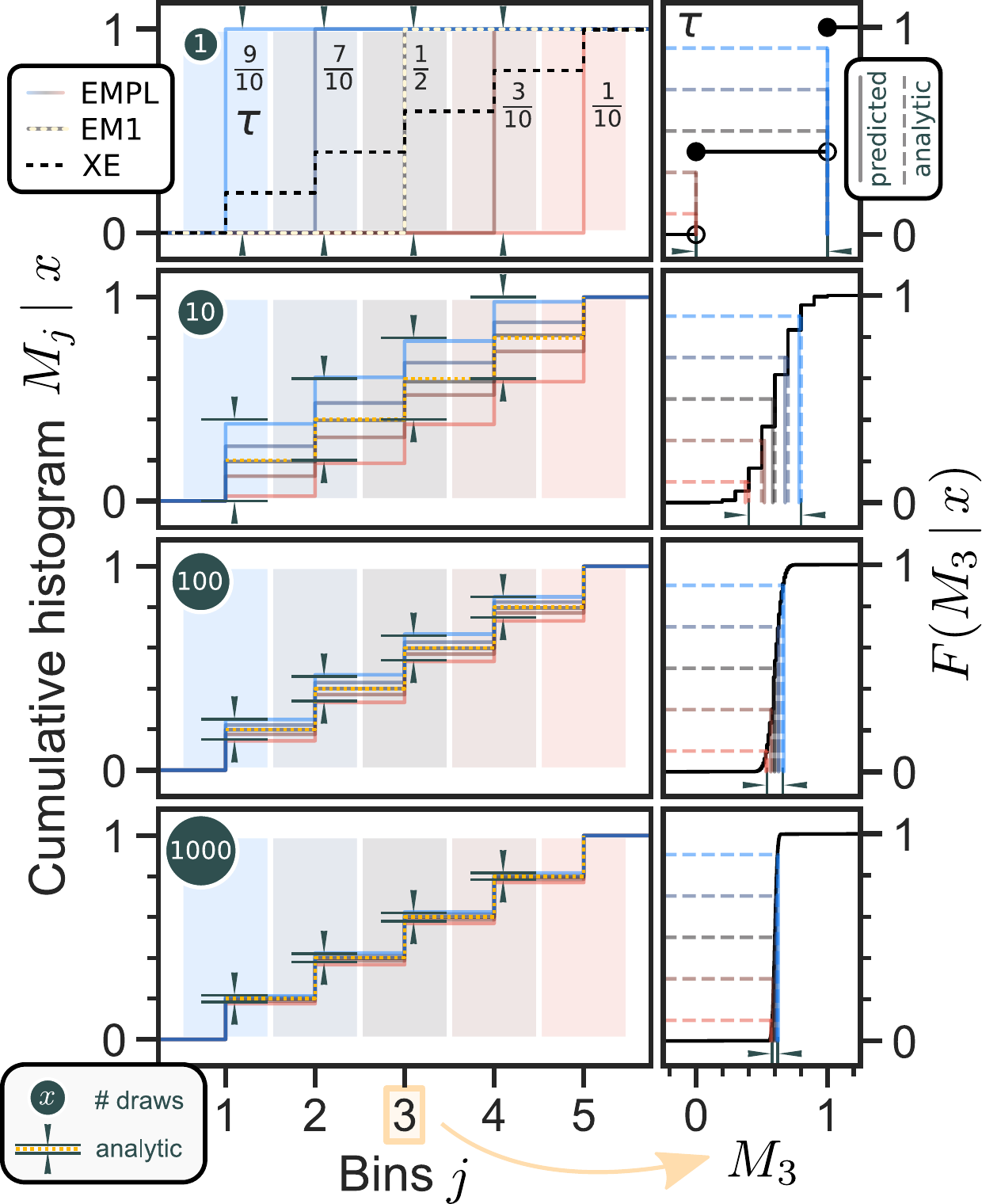}
    }
    \vspace{-0.2cm}  
    \caption{
    \emph{Left}: EMPL predictions for the toy example, for $x = $ 1, 10, 100, and 1,000 draws. Each colored line corresponds to the MLP prediction for a particular quantile level $\tau$ (1/10, $\ldots$, 9/10), specified next to it in the top panel. The gray markers delimit the true inter-quantile range between the lowest and highest considered quantile levels, and the true median is indicated by the golden dotted line. For a single draw $x = 1$, the predictions of NNs trained using the 1-Wasserstein distance (EM1) and the cross-entropy loss (XE; computed w.r.t. to the density histogram $(m_j)_{j=1}^N$) are also shown: both produce a correct ``average histogram'' in a certain sense (namely median and mean, respectively), but a single number per bin is not sufficient  to properly reflect the \emph{distribution} of possible histograms. \emph{Right}: Analytic CDF of the cumulative histogram $M_3$ in the central bin $j = 3$ (solid black line). The predicted quantiles (solid vertical lines) agree with the analytic quantiles (dashed vertical lines) for all values of $x$ and quantile levels $\tau$ (dashed horizontal lines). As $x \to \infty$, all the quantiles converge towards $j / N = 3 / 5$.
    }
    \label{fig:toy_example}
    \vspace*{-0.3cm}
\end{figure}

We now present histogram regression results from three experiments with the EMPL. We start with a toy example intended to build some intuition for the problem at hand, also showing that predicting average histograms with standard loss functions is insufficient when the data exhibits high stochasticity. Then, we consider an application to sports and lastly, we study a computer vision task in $\gamma$-ray astronomy. An additional example that considers a bimodal distribution within each bin is provided in the Supplementary Material.

\subsection{Toy example: drawing balls from an urn}
\label{sec:toy}
First, we illustrate our method by means of a toy example, for which the analytic solution can be computed. Minimizing the pinball loss is equivalent to maximizing the likelihood of an asymmetric Laplace distribution \citep{Yu2001}, which requires the outcome in each bin to be continuous. Whilst interpolation techniques such as \emph{jittering} could be applied to the outcome in the discrete case (and are needed in fact to obtain analytical convergence results; \citealt{Machado2005, Padellini2018}), we will show in this example that even in the extreme case where the set of possible outcomes consists of the two integers $\{0, 1\}$, the predictions for the quantiles \emph{in practice} behave as expected.
\par The scenario is the following: we randomly draw $x$ times with replacement from an urn that contains numbered but otherwise identical balls $\mathcal{N} = \{1, \ldots, N\}$ such that each ball has equal probability of being drawn. We keep track of the drawn numbers by adding a tally mark in the respective field (or bin) of a table after each draw before putting the ball back into the urn again. This yields a frequency histogram $(m_j^c)_{j=1}^N$, with $m_j^c$ denoting the number of times the ball $j$ has been drawn. The task of the NN will be to estimate the distribution of the \emph{relative} counts in each bin $(m_j)_{j=1}^N$, where $m_j = m_j^c / x$, depending on the number of draws $x$. 
\par The total number of counts $m_j^c$ in each bin $j \in \mathcal{N}$ follows a binomial distribution $B(x, p)$, where $p = 1 / N$. Since the mean and variance of the relative counts in each bin are given by $\mathbb{E}[m_j] = p$ and $\text{Var}(m_j) = p (1 - p) / x$, the relative counts $m_j \to p$ as $x \to \infty$, for all $j \in \mathcal{N}$, implying that each ball will be drawn equally often in the (hypothetical) limit of infinitely many draws. However, for small values of $x$, the variability of the resulting histograms is high, and in the case of a single draw $x = 1$, $m_j = 0$ in $N - 1$ bins, while $m_j = 1$ in the bin for the drawn number $j$. 
\par Since the EMPL compares the \emph{cumulative} histograms, we also define the cumulative histogram as $(M_j^c)_{j = 1}^{N}$, where $M_j^c = \sum_{r = 1}^j m_r^c$, and similarly for the \emph{relative} cumulative histogram $(M_j)_{j=1}^N$. We write $Y \sim \mathcal{U}\{1, N\}$ for the random variable $Y$ describing a single draw from the urn (i.e. from a discrete uniform distribution between $1$ and $N$).
We can determine the CDF for the cumulative counts $M_j^c$ in each bin $j \in \mathcal{N}$ by computing the conditional probability for drawing at most $l \in \{0, \ldots, x\}$ times a number less than or equal $y \in [1, N]$, given by
\begin{equation}
\begin{aligned}
    P(\#(Y \leq y) \leq l \, | \, x) = P(M_j^c \leq l \, | \, x) \\
    = \sum_{m = 0}^l {x \choose m} p_{{\scriptscriptstyle\leq} j}^m \left(1 - p_{{\scriptscriptstyle\leq} j}\right)^{x - m},
\end{aligned}
\end{equation}
where $j = \left \lfloor{y} \right \rfloor$ (only integers can be drawn), $p_{{\scriptscriptstyle\leq} j} = j \, / \, N$ is the probability for drawing a number less than or equal $y$ in a \emph{single} draw, and the probabilities for drawing \emph{exactly} $0, \ldots, l$ times a number $Y \leq y$ need to be summed up to obtain the probability for drawing \emph{at most} $l$ times such a number.
Inverting this relation yields the quantiles for the distribution of the value $M_j^c$ (and equivalently $M_j$) in each bin, conditional on $x$.
\par For our numerical experiment, we choose $N = 5$ balls and take the number of draws $x$ itself to be a random variable $X$, given by $X = \operatorname{round}(\hat{X})$ with $\log_{10}(\hat{X}) \sim U(0, 3)$. We train a simple multilayer perceptron (MLP) containing 2 hidden layers with 128 neurons each, ReLU activation and batch normalization for the hidden layers, and a softmax activation for the output layer to obtain $\tilde{Q}^\theta(X; \tau)$ as described in Sec.~\ref{subsec:implementation}.
The NN training consists of 10,000 batch iterations at a batch size of 2,048. We minimize the EMPL with randomly drawn quantile levels $\tau$ using an Adam optimizer \citep{Kingma2014}. The 2-dimensional inputs to the NN are given by $x$ and $\tau$, and the corresponding 5-dimensional labels $(m_j)_{j=1}^N \, | \, X = x$ are generated by randomly drawing $x$ times from a discrete uniform distribution with range $\mathcal{N}$ and normalizing the resulting histogram. Equivalently (and faster), one can draw from a multinomial distribution with $x$ trials and uniform probability $p_j = p = 1/N$ for all $j \in \mathcal{N}$.
\par Fig.~\ref{fig:toy_example} shows the quantiles of the estimated cumulative relative counts in each histogram bin for $x = $ 1, 10, 100, and 1,000 draws (left panels). The colored lines correspond to the predicted quantiles as indicated in the top panel next to the lines, and the dark gray delimiters show the analytic values for the two most extreme considered quantiles ($\tau = 1/10$ and $9/10$). The right panels depict the true CDFs of the relative cumulative counts in the central bin for each $x$, i.e. $F(M_3 \, | \, x)$, together with the analytic (dashed) and predicted (solid) quantiles, given by the pre-image of $\tau$ (see the horizontal dashed lines) under the CDF.
\par For a single draw from the urn, i.e. $x = 1$, the only possible values of the histograms are 0 and 1. The cumulative histogram in bin $j$ for a quantile level $\tau$ should be 1 if $\tau$ is greater than the probability of drawing a number greater than $j$, i.e. if $\tau > (N - j) / N$, and 0 else. For all values of $\tau$, the MLP correctly determines where the cumulative histogram jumps from 0 to 1. Since the EMD coincides with the EMPL for $\tau = 0.5$, a NN trained by minimizing the EMD predicts $m_j = 1$ for the central bin $j = 3$ and $m_j = 0$ otherwise.\footnote{See the Supplementary Material for the expected EMD for $x = 1$.}
In contrast, using the cross-entropy loss for training produces the expected \emph{mean} histogram in each bin ($m_j = 1/N$; independently of the input $x$), which for $x = 1$ is not representative of any observed histogram with values in $\{0, 1\}$. Therefore, among the considered loss functions, only the EMPL is able to adequately express the full \emph{range} of possible histograms, thanks to the dependence of the NN outputs on the quantile level of interest $\tau$.
\par As the number of draws $x$ increases, the CDF of $M_j$ gradually becomes narrower (see the right panels for the central bin $j = 3$) and consequently, the quantile ranges converge from both sides towards $p_{{\scriptscriptstyle\leq} j} = j / N$. For all $x$ and $\tau$, the predicted quantiles match their analytic counterparts. Note that in the limit of no stochasticity $x \to \infty$, the NN prediction with the EMPL would become $\tau$-independent and equal its cross-entropy and EMD loss counterparts. 
\begin{figure}[!t]
    \centering
    \resizebox{0.88\columnwidth}{!}{
    \includegraphics{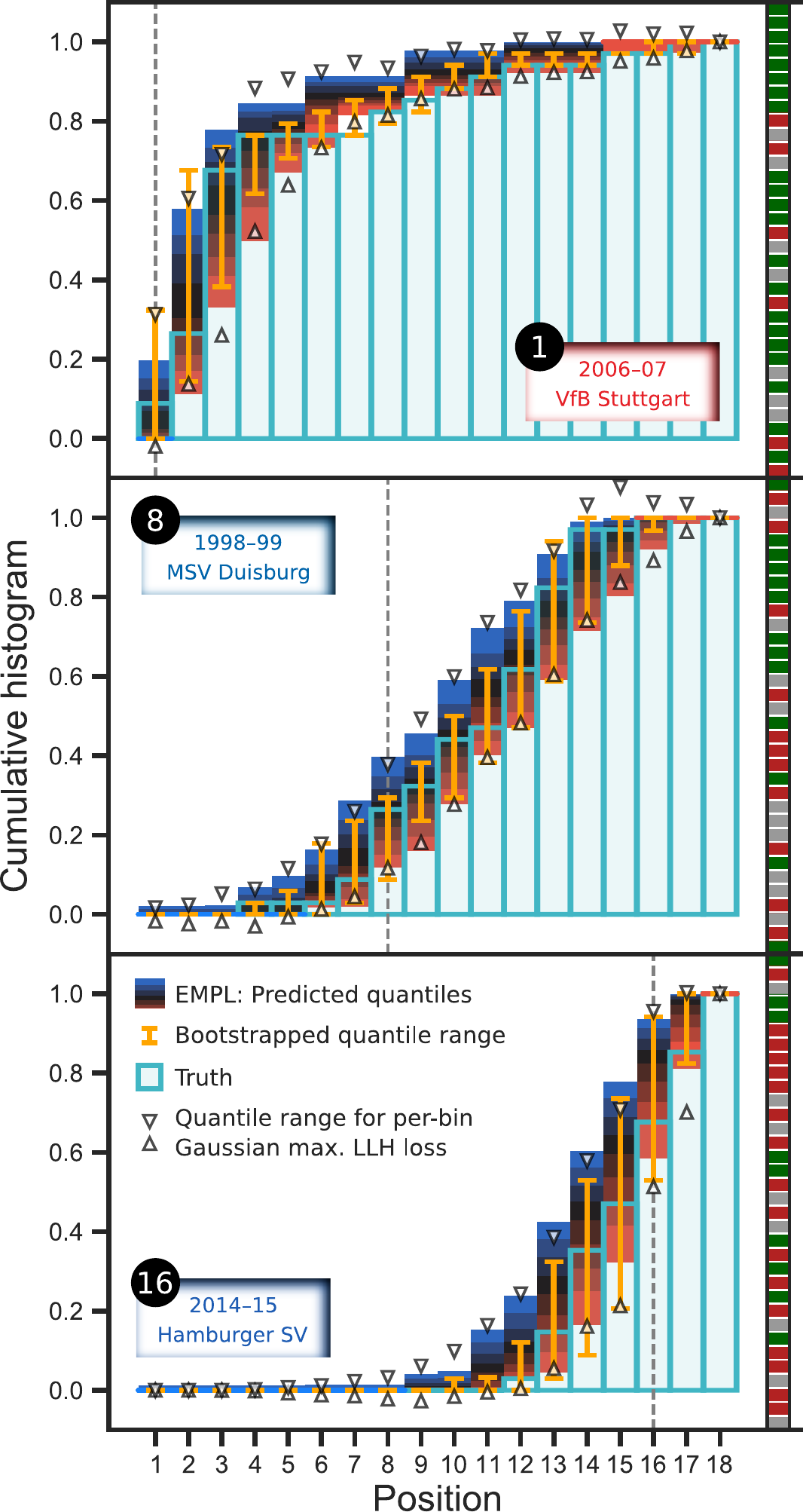}
    }
    \caption{
    Cumulative histograms depicting the distribution of the table position after each week for three clubs (one from each test season). The light-blue histograms represent the truth, and the colored regions show the estimated quantiles from $10 - 90\%$ in steps of $10\%$ with the EMPL. The orange error bars show bootstrapping estimates of the $10 - 90\%$ quantile range. When simply training a NN to estimate bin-wise means and standard deviations for the cumulative histogram assuming a Gaussian likelihood (independently in each bin), the resulting quantile range is not confined to $[0, 1]$ due to the infinite support of the Gaussian distribution and may be non-monotonic (white triangles), which are undesirable properties.
    The vertical dashed lines and white numbers indicate the position of the club at the end of the respective season (which cannot be inferred from the histograms). The results of all the club's matches during the season (i.e., the input $X$ for the MLP) are illustrated to the right of the histograms (see main text).
    }
    \label{fig:bundesliga}
\vspace*{-0.3cm}
\end{figure}
\subsection{An application to the football Bundesliga}
Week after week, millions of fans around the globe cheer passionately for their favorite football club in the hope of claiming the league title by finishing first at the end of the season. Whilst each club tries its best to win as many matches as possible, their fortune is not entirely in their own hands: 38 points at the end of the Bundesliga (German top-flight division) season 1997$-$98 were not enough to save Karlsruher SC from being relegated to 2. Bundesliga placed 16th; however, the same number of points would have sufficed for position 13 in season 2001$-$02, in safe distance from the relegation spots 16$-$18. Thus, knowledge of the results of a single club \emph{in isolation} is a strong indicator of how well the club fares in terms of the league table, but is not sufficient to determine its position.
\par We apply the EMPL to the following task: given the list of points that a club has earned in each match during a season $X \in \{0, 1, 3\}^{34}$  (win: 3 points, draw: 1 point, defeat: 0 points; for 34 matches), estimate the histogram $(m_j)_{j=1}^N$ that results from the positions of the club in the league table after each week. Narrow histograms (steep cumulative histograms) indicate few change in the position over the course of the season, while a wide histogram (a gently increasing cumulative histogram) suggests a turbulent season for the respective club in terms of its place in the table. For instance, if a club managed to lead the table throughout the season over 34 weeks, this would result in a histogram with $m_1 = 1$ and $m_j = 0$ for $j = 2, \ldots, 18$ (where 18 is the number of competing clubs).
\par We use data from all the Bundesliga seasons between 1995$-$96 (when the 3-points-for-a-win rule was introduced) and 2017$-$18, keeping the seasons 1998$-$99, 2006$-$07, and 2014$-$15 for testing, while using the other 20 seasons as training data.\footnote{{\scriptsize Data: \url{www.kaggle.com/thefc17/bundesliga-results-19932018}}.} In order to increase the amount of training data, we ``re-play'' each training season 1,000 times, randomly permuting the 34 weeks (each of the 18 clubs plays $2 \times 17 = 34$ matches in a season, namely a home and an away match against every other club). Clearly, these artificial seasons converge to the same league table by the end of the season, but the histograms of the table positions after each week are distinct (as is the ordering of the input lists $X$ containing the points from each match).
\par We train again a simple MLP with 2 hidden layers with 128 neurons followed by ReLUs and 50\% dropouts \citep{Hinton2012, Srivastava2014} to prevent overfitting. A softmax activation function produces the relative histograms, consisting of 18 bins corresponding to the league table positions. We minimize the Smoothed EMPL with $\alpha = 0.005$ over 250 epochs using a batch size of 2,048.
\par Fig.~\ref{fig:bundesliga} shows the predictions for the relative cumulative histograms for three clubs (one from each test season). The light-blue histogram corresponds to the true cumulative histogram for the club in the respective season, and the gray dashed line shows the final standing of the club (which cannot be deduced from the histogram). 
The colored regions indicate the estimated quantile ranges, from 10 to 90\% in steps of 10\%. The orange error bars show bootstrapping estimates of the bin-wise quantiles obtained by computing hypothetical histograms that would arise had the club obtained the same points in each match in \emph{another} (artificially generated) season. Specifically, we randomly select 200 seasons from our augmented dataset, remove the club whose final number of points is closest (in order to minimize the bias due to situations such as having two champions from different seasons compete against each other, which would bias the histograms towards lower positions), and calculate the bin-wise quantiles over the resulting histograms. The lists of points in each match (i.e. the inputs $X$ fed to the MLP) are illustrated on the right-hand side next to the histograms (1st match at the bottom, 34th match on top; green: 3/win, gray: 1/draw, red: 0/defeat). Although the estimated quantile ranges do not perfectly match their bootstrapping counterparts, their magnitudes are generally similar, and the EMPL enables the quantification of the uncertainty in the distribution of a club's table position over the season, based on other seasons and without any knowledge about the results of the other clubs. Thus, the available domain knowledge that is learned during the training in combination with a limited number of observations allows one to derive a narrow posterior distribution that expresses which histograms are compatible with the observations.
\par A clear advantage of quantile-based approaches is that they do not assume a specific underlying distribution. We illustrate this by comparing our method with a naive likelihood-based approach for quantifying the bin-wise uncertainty in the histograms, namely a NN trained by simply maximizing the Gaussian log-likelihood for the cumulative histogram $(M_j)_{j=1}^N$ with independent means $\mu_j$ and standard deviations $\sigma_j$ for each bin (resulting in $2 \times 18 = 36$ outputs). While the monotonicity of the mean estimates $(\tilde{\mu}_j)_{j=1}^N$ and the normalization $\tilde{\mu}_N = 1$ are enforced by computing $\tilde{\mu}$ as the cumulative sum of softmax-activated logits (see Sec.~\ref{subsec:implementation}), other quantiles of the cumulative histograms do not need to be monotonic (caused by $\tilde{\sigma}_j \neq \tilde{\sigma}_{j-1}$); also, the values are unbounded due to the infinite support of the Gaussian distribution. In contrast, the EMPL predictions, which do not presume a particular distribution of the histogram values within each bin, lie in $[0, 1]$ and increase monotonically for all $\tau$ \emph{by construction}, which are essential properties.

\subsection{An example from astrophysics: predicting point-source brightness distributions}
\begin{figure}[tb]
    \centering
    \resizebox{1\columnwidth}{!}{
    \includegraphics{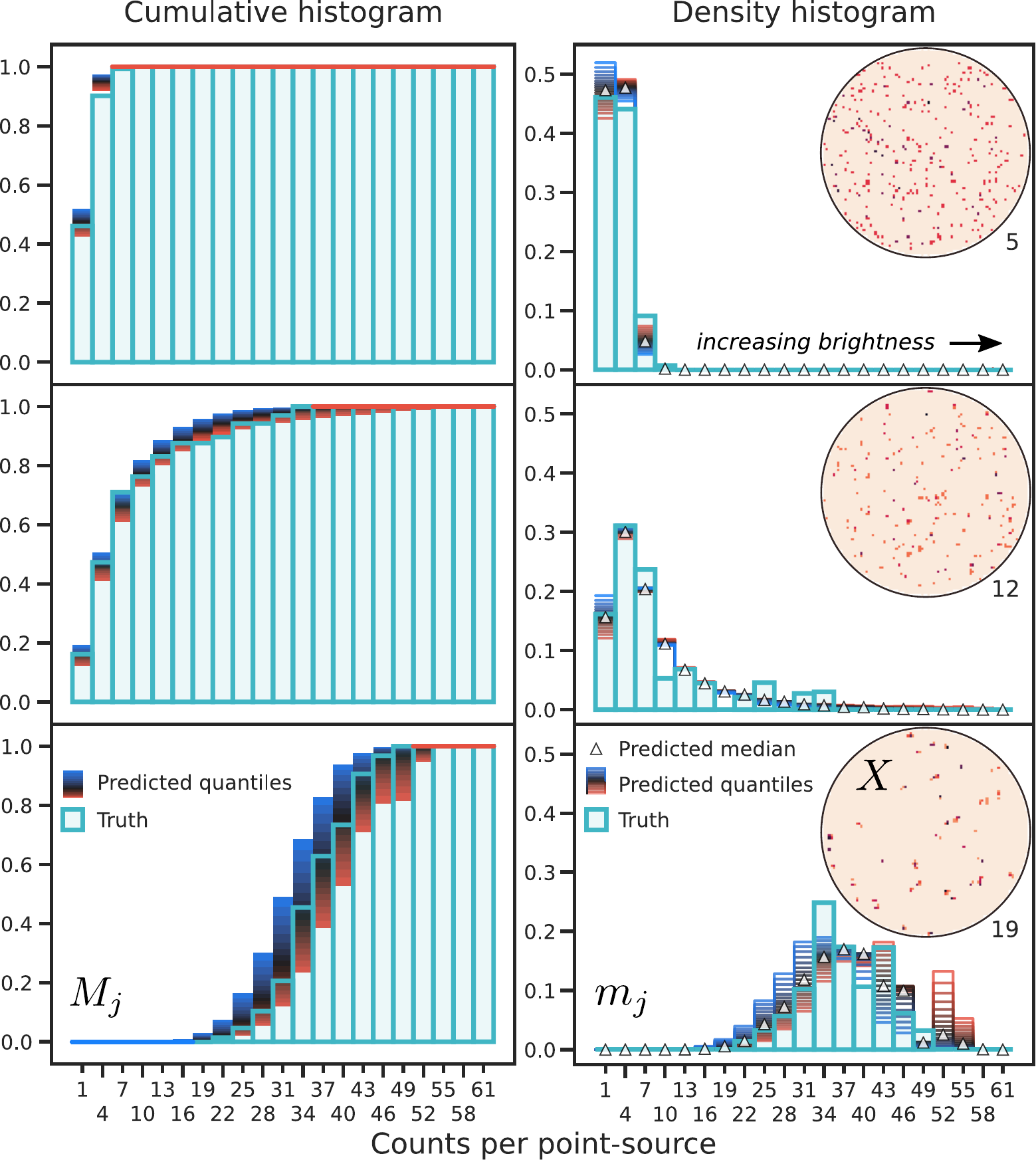}
    }
    \vspace{-0.3cm}
    \caption{
    Predicted and true cumulative (left) and resulting density (right) point-source brightness histograms. Projections of the corresponding input photon-count maps $X$ are shown in the inset plots (log-scaled), and the maximum number of counts per pixel (which is considerably lower than the maximum number of counts \emph{per point-source} because of the point spread function), which defines the upper limit of the color map, is indicated next to the maps. 
    }
    \label{fig:gamma_ray}
\vspace*{-0.3cm}
\end{figure}
\addtolength{\tabcolsep}{-2pt} 
\begin{table}[t]
\begin{tabular}{llllll}
\toprule
\begin{tabular}[c]{@{}l@{}}Training loss\end{tabular} &
  \begin{tabular}[c]{@{}l@{}}MAE$^{1}$\\ \small{$\downarrow$}\end{tabular} &
  \begin{tabular}[c]{@{}l@{}}MSE$^1$\\ \small{$\downarrow$}\end{tabular} &
  \begin{tabular}[c]{@{}l@{}}EM1$^2$\\ \small{$\downarrow$}\end{tabular} &
  \begin{tabular}[c]{@{}l@{}}EM2$^2$\\ \small{$\downarrow$}\end{tabular} &
  \begin{tabular}[c]{@{}l@{}}IS [\%]\\ \small{$\uparrow$}\end{tabular} \\ \midrule
MAE                      & 4.1 & 8.6 & 10.8 & 4.6 & 95.7 \\
MSE                      & 4.0 & \textbf{8.1} & 10.3 & 4.2 & 95.8 \\
XE                       & 4.2 & 8.6 & 9.4  & 4.2 & 95.6 \\ \cdashlinelr{1-6}
EMP (\textbf{all} $\tau$)  & 4.0 & 8.3 & 8.8  & \textbf{3.9} & 95.8 \\
SEMP (\textbf{all} $\tau$) & 4.0 & 8.2 & 9.0  & \textbf{3.9} & 95.8 \\
EM1 ($\tau = 0.5$)       & \textbf{3.9} & 8.2 & \textbf{8.7}  & \textbf{3.9} & \textbf{95.9} \\ \bottomrule
\end{tabular}
\footnotesize{$^1$: $\times$ 1,000, \hspace{0.5cm} $^2$: $\times$ 100 \\
MAE / MSE: mean absolute / squared error, XE: cross-entropy, \\
(S)EMP: (Smoothed) EMPL (always \emph{evaluated} for $\tau = 0.5$), $\alpha = 0.001$ when smoothed, EM1/2: absolute/squared EMD, \\
IS: histogram intersection \cite{Swain1991}.}
\vspace*{-0.2cm}
\caption{
Different metrics (columns) when evaluating the NN on 512 test maps, for NNs trained using different loss functions (rows). The EM-based losses perform similarly to the per-bin losses in terms of per-bin metrics (MAE/MSE/IS), and achieve better results as measured by the EMD. Training the NN for all quantile levels $\tau$ simultaneously barely affects the median accuracy as compared to the EM1 loss ($\tau = 0.5$ only), while yielding much more expressive outputs through \emph{arbitrary} quantiles, thus providing \emph{uncertainties}.}
\label{tab:gamma_ray_table}
\vspace{-0.3cm}
\end{table}
\addtolength{\tabcolsep}{2pt}
Now, we consider a computer vision problem in the field of $\gamma$-ray astronomy, namely the estimation of the point-source brightness distribution $(m_j)_{j=1}^N$ given a photon-count map $X \in \mathbb{N}^{n_\text{pix}}$ as an input, where $n_\text{pix}$ is the number of pixels in the region of interest (ROI, taken to be a circle of radius $20 \degree$ here). Specifically, each observed map $X_s$ contains emission from $T_s \in \mathbb{N}$ point-sources, each of which contributes $C_s^t\in \mathbb{N}$ photon counts to the map (for $t = 1, \ldots, T_s$), such that the total number of counts in map $s$ is $C^\text{tot}_s = \sum_{t=1}^{T_s} C_s^t$. Binning the counts in the map according to $(C_s^t)_{t=1}^{T_s}$ results in a histogram that characterizes the brightness distribution of the generating point-source population: for each $t = 1, \ldots, T_s$, $C_s^t$ counts are added to the associated bin (for example, for a source $t$ responsible for $C_s^t = 4$ counts, these 4 counts are added to the ``3$-$5 counts'' bin), implying that the counts from dim (bright) point-sources go to low (high) bins. Once all the counts are distributed, the histogram is normalized to sum up to unity. The task of the NN is to estimate this underlying point-source brightness distribution $(m_j)_{j=1}^N$ from a photon-count map $X$.
\par A particularly interesting application is the analysis of the photon-count map from the \emph{Fermi} space telescope \citep{Abdollahi2020}, which contains unexplained excess emission from the center of our Milky Way galaxy \citep{Goodenough2009} that could possibly be explained by annihilation of dark matter particles (\citealt{Hooper2011}). Recently, machine learning methods have opened up a new avenue for the analysis of this excess \citep{Caron2018, List2020b, Mishra-Sharma2020}. 
Whilst an exhaustive study of the \emph{Fermi} map is beyond the scope of this work, we demonstrate that our method is able to estimate the histogram describing the brightness distribution of point-sources in a simple scenario with simulated photon-count maps.
We generate 312,500 photon-count maps with the tool \texttt{NPTFit-Sim} \citep{NPTFit-Sim}, modeling emission from isotropically distributed point-sources. The photon counts from each source are smeared out by the \emph{Fermi} point spread function over multiple pixels, and each pixel may contain counts from more than one source, making the problem probabilistic and non-trivial. We subsequently discard the maps with less than 1,000 counts and those that contain very bright point-sources with $> 60$ counts; then, we put aside 1/15th of the remaining maps for testing and use the others as training data.
\par The input maps $X_s$ are discretized using the \texttt{HEALPix} tessellation of the sphere \citep{Gorski2005}, and we use a resolution set by the parameter $N_\text{side} = 256$ (giving $n_\text{pix} = $ 65,536 in our ROI).
As proposed by \citet{List2020b}, we employ a graph-convolutional NN built on the DeepSphere framework \citep{Perraudin2019, Defferrard2020}, in which the \texttt{HEALPix} sphere is described by a weighted undirected graph, and the convolution operation is defined by means of the graph Laplacian operator. Our NN is composed of 8 graph-convolutional layers, each followed by maximum pooling, batch normalization, and a ReLU activation, and three fully-connected (FC) layers. The quantile level $\tau$ is appended before the first FC layer.
\par Fig.~\ref{fig:gamma_ray} shows three examples from the testing dataset, for a dim (top), moderate (middle), and bright (bottom) point-source population. The cumulative and density histograms are depicted in light blue (truth) and by colored regions / lines (NN), corresponding to $5 - 95 \%$ quantiles in steps of $5\%$.
The white triangles in the right panels are located at the predicted medians. The NN has learned to faithfully recover the underlying brightness histograms. The uncertainties become larger with increasing brightness for equally spaced bins as considered here; however, we found in our experiments that this trend generally reverses when using logarithmically spaced bins.
\par Table~\ref{tab:gamma_ray_table} lists several metrics when using different loss functions for the NN training, evaluated on 512 testing maps.
We emphasize that in the case of the median $\tau = 0.5$, for which we report our results with the EMPL, the EMPL by construction is \emph{exactly identical} to the EMD (see Eq.~\eqref{eq:EMPL_generalises_EMD}, up to $1/2$), as the EMPL naturally extends the EMD to arbitrary quantiles. Therefore, one should not expect a higher accuracy when evaluating the EMPL-trained NN for the specific value $\tau = 0.5$ as compared to the EMD-trained NN. In turn, Table~\ref{tab:gamma_ray_table} shows that training the same NN to estimate \emph{all} the quantiles using the EMPL rather than only the median \emph{barely affects} the accuracy of the median predictions. Thus, the EMPL provides much more expressive outputs that quantify the uncertainties ``for free''. The EM-based losses (EM1 \& EMPL) outperform the bin-wise losses w.r.t. the cross-bin metrics EM1 and EM2, while performing similarly in terms of the bin-wise metrics.
\par Whilst labeled histogram data may be difficult to acquire or might not be available at all in some applications, this astrophysical example belongs to the important class of problems where labeled training data can be obtained (for instance using a simulator), but \emph{recovering} the underlying histogram from real data is a challenging task that can be tackled by Deep Learning methods. For instance, CNNs are able to assess the real photon-count map of the sky on multiple scales, which can potentially give rise to more robust results in the presence of mismodeling on large angular scales \citep{List2020b}, whereas statistical methods typically rely on an approximation of the likelihood that treats each pixel independently (e.g. \citealt{Mishra-Sharma2017}).

\section{Conclusions}
\label{sec:conclusions}
We have presented a method for the NN-based regression of histograms from input images or other data. Our approach is based on minimizing a novel loss function, which we call the \emph{Earth Mover's Pinball Loss} (EMPL), rooted in transportation theory as well as in quantile regression. This loss function is an asymmetric generalization of the EMD that allows for the regression of \emph{arbitrary quantiles} of the cumulative histogram in each bin, harnessing the idea of the pinball loss.
In the particular case of the median ($\tau = 0.5$), our loss function reduces to the EMD. We have demonstrated the effectiveness of our method in a toy example, a football-related task, and a problem in $\gamma$-ray astronomy. The accuracy of the estimated median histogram
is very similar to the standard EMD case, and the prediction of arbitrary other quantiles comes at almost no additional cost (the increase in walltime for training is $<$ 10\%). Given the vast range of applications where histograms are used, there is a great potential for Deep Learning methods to provide accurate, fast, and reliable histogram predictions. The EMPL is easy to implement (see the Supplementary Material), and we expect it to be particularly useful for tasks where the entire \emph{distribution} of possible histograms is of interest such as rain forecasts (``what's the probability that it rains more than 10 mm tomorrow?'') or radiation treatment planning (``how certain can we be that 20\% of the cancerous organ should receive a dose of 30 Gy?''). Possible extensions of our work include multidimensional histograms, incorporating epistemic uncertainties (e.g. \citealt{Gal2016}), flexible ground distances, and the application to parameterized continuous (i.e. unbinned) distributions.

\section*{Acknowledgments}
The author is grateful to G. F. Lewis for his support and helpful discussions. Also, the author wishes to thank N. Rodd for his useful feedback and C. Proissl for suggesting a simplification for the toy example. The author acknowledges the National Computational Infrastructure (NCI), which is supported by the Australian Government, for providing services and computational resources on the supercomputer Gadi that have contributed to the research results reported within this paper. The author is supported by the University of Sydney International Scholarship (USydIS).\\[0.2cm]
\emph{Software}: \texttt{matplotlib} \cite{mpl}, \texttt{seaborn} \cite{seaborn}, \texttt{numpy} \cite{npy}, \texttt{scipy} \cite{scipy}, \texttt{numba} \cite{numba}, \texttt{healpy} \cite{healpy}, \texttt{Tensorflow} \cite{Abadi2016}, \texttt{Keras} \cite{keras}, \texttt{ray} \cite{ray}, \texttt{dill} \cite{dill}, \texttt{cloudpickle},\footnote{\href{https://github.com/cloudpipe/cloudpickle}{https://github.com/cloudpipe/cloudpickle}} \texttt{colorcet}.\footnote{\href{https://github.com/holoviz/colorcet}{https://github.com/holoviz/colorcet}} Also, we used the free software \texttt{Inkscape}.\footnote{\href{https://inkscape.org/}{https://inkscape.org/}}

\clearpage
\renewcommand{\thefigure}{S\arabic{figure}}
\renewcommand{\thesection}{S\arabic{section}}
\renewcommand{\theequation}{S\arabic{equation}}
\setcounter{figure}{0}
\setcounter{section}{0}
\setcounter{equation}{0}
\onecolumn
\vspace*{-0.5cm}
\icmltitle{\textsc{Supplementary Material} \\[0.2cm]
The Earth Mover's Pinball Loss: \\ Quantiles for Histogram-Valued Regression \\[0.2cm]
\normalsize{Florian List}}
\vskip 0.3in
\section{Expected EMD in the Toy Example for a Single Draw}
\begin{figure}[!htb]
    \centering
    \resizebox{0.85\textwidth}{!}{
    \includegraphics{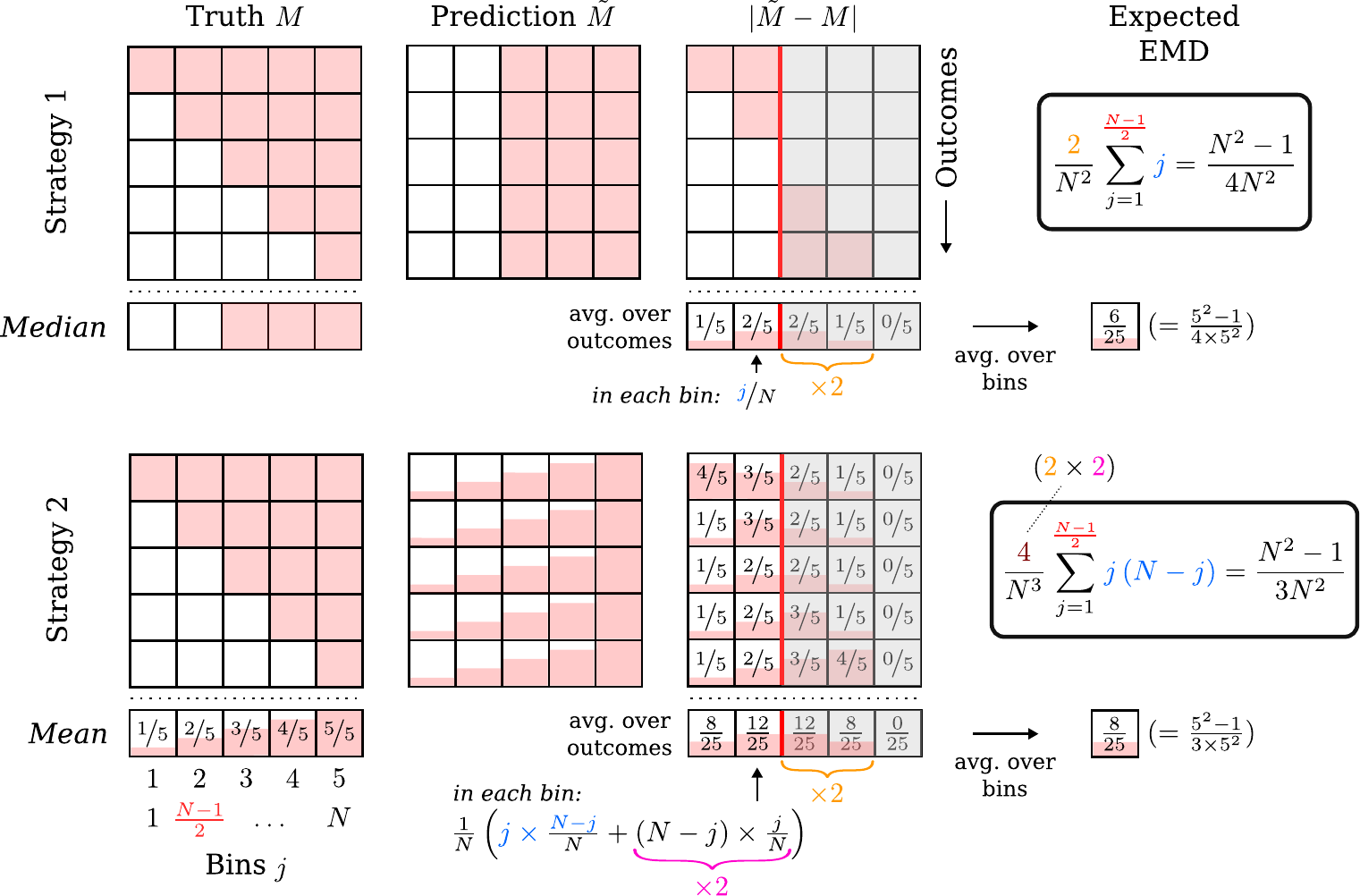}
    }
    \caption{
    Sketch illustrating the calculation of the expected EMD with Strategies~1 and 2, where the predicted cumulative histogram $\tilde{M}$ is given by the \emph{median} and the \emph{mean} over all possible cumulative histograms $M$, respectively. We show the case $N = 5$ as considered in the main body. Within each grid, the columns correspond to the bins $j \in \{1, \ldots, N\}$, and each row belongs to a possible outcome $Y \in \{1, \ldots, N\}$. The filling of each square indicates the value, from 0 (empty) to 1 (completely filled). 
    The rightmost grids show the absolute difference $|\tilde{M} - M|$ between the predicted and true cumulative histogram for each outcome and bin. The expected EMD for each strategy is given by the mean of $|\tilde{M} - M|$ over the outcomes (vertically) and over the bins (horizontally), and can therefore be read off as the filled area fraction (e.g., 6 out of 25 squares are filled with Strategy 1 for $N = 5$). Due to the symmetry of $|\tilde{M} - M|$ w.r.t. the central bin, the total filled areas to the right and to the left of the red vertical line are equal, for which reason it is sufficient to consider $j \in \{1, \ldots, (N - 1) / 2\}$ and to multiply the result by two (in yellow) in the calculation of the expected EMD. With Strategy~1, averaging $|\tilde{M}_j - M_j|$ over the outcomes gives $j / N$ for bins $j \leq (N - 1) / 2$, whereas one obtains $1/N \big{(}j (N - j)/N + (N - j) j/N \big{)}$ with Strategy~2, for example $\frac{12}{25} = \frac{1}{5} \left(2 \times \nicefrac{3}{5} + 3 \times \nicefrac{2}{5}\right)$ for $j = 2$ and $N = 5$.
    }
    \label{fig:EMD_proof_sketch}
\end{figure}
In this section, we briefly revisit the toy example from the main body (Sec.~\ref{sec:toy}) and discuss the minimization of the expected EMD for a single draw. Recall that the experiment consists of $x$ times randomly drawing a numbered ball from $\mathcal{N} = \{1, \ldots, N\}$ with replacement, where each draw is described by a random variable $Y \sim \mathcal{U}\{1, N\}$ that follows a discrete uniform distribution. The histogram label $m = (m_j)_{j=1}^N$ in bin $j$ expresses the fraction of times the number $j$ was drawn (that is, $m_j = 0$ if number $j$ was never drawn, and $m_j = 1$ if $j$ was drawn $x$ times). For a \emph{single} draw, i.e. $x = 1$, we noted that a histogram estimate that places all probability mass in the central bin minimizes the expected EMD towards the true outcome. Therefore, this is what a NN trained using the EMD (and hence an EMPL-trained NN just as well for $\tau = 0.5$) predicts in this case, as can be seen in the top panel in Fig.~\ref{fig:toy_example}.
Here, we calculate the expected EMD as a function of $N$ for this strategy and compare it with the case of a histogram estimate that uniformly distributes the probability mass over the bins. We restrict ourselves to the case of odd $N$, such that there is a unique central bin. In the discrete case of normalized histograms in 1D, the EMD (or 1-Wasserstein distance) between predicted histogram $\tilde{m}$ and observed realization $m$ is given by
\begin{equation}
    W_1(\tilde{m}, m) = \frac{1}{N} \sum_{j=1}^N |\tilde{M}_j - M_j|,
\end{equation}
where $M_j = \sum_{r=1}^j m_r$ denotes the observed cumulative histogram, and similarly for $\tilde{M}_j$. Note that in order not to overload notation, we do not distinguish between the random variables $m = (m_j)_{j=1}^N$ (and $M = (M_j)_{j=1}^N$) that express the random histogram values in each bin and realizations of these random variables in this Supplementary Material.

\par \textbf{Strategy 1} (median): \emph{Putting everything on} $(N + 1) / 2$ \\
In this strategy, the estimated histogram $\tilde{m}^1 = (\tilde{m}^1_j)_{j=1}^N$ is $\tilde{m}^1_j = 1$ for $j = (N + 1) / 2$ (the central bin) and $\tilde{m}^1_j~=~0$ otherwise. This choice minimizes the expected EMD, which immediately follows from the fact that the associated \emph{cumulative} histogram $\tilde{M}^1_j = 0$ for $j < (N + 1) / 2$ and $\tilde{M}^1_j = 1$ for $j \geq (N + 1) / 2$ is the \emph{median} over the possible cumulative outcomes in each bin (see the top left grid in Fig.~\ref{fig:EMD_proof_sketch}), recalling that the median minimizes the mean absolute error. To obtain the expected EMD, we will first compute the mean of $|\tilde{M}^1_j - M^1_j|$ over the $N$ possible outcomes ($Y = 1, \ldots, N$) for each bin $j$, and then take the mean over the bins. Since the possible outcomes as well as the estimated histogram are symmetric w.r.t. the central bin, it is sufficient to consider $j \in \left\{1, \ldots, (N - 1)/2\right\}$ (i.e., the bins to the left of the red vertical line in the rightmost grids in Fig.~\ref{fig:EMD_proof_sketch}), for which one finds that the expected absolute difference between the estimated and true cumulative histogram is given by
\begin{equation}
    \mathbb{E}_m\left[|\tilde{M}^1_j - M_j|\right] = 
        \frac{j}{N}.
\end{equation}
Accounting for the symmetry and averaging over the bins $j = 1, \ldots, N$ yields
\begin{equation}
    \mathbb{E}_m\left[W_1(\tilde{m}^1, m)\right] = \frac{2}{N^2} \sum_{j=1}^{\frac{N-1}{2}} j = \frac{N^2 - 1}{4 N^2}.
\end{equation}

\par \textbf{Strategy 2} (mean): \emph{Uniformly distributing the probability mass} \\
For comparison, we consider another possible strategy, where the probability mass is uniformly distributed over the bins, and the estimated histogram $\tilde{m}^2 = (\tilde{m}^2_j)_{j=1}^N$ is defined by $\tilde{m}^2_j = 1/N$ (and hence $\tilde{M}^2_j = j/N$). Note that this choice corresponds to the \emph{mean} over all possible realizations of the cumulative histogram. As above for Strategy 1, we compute the expected absolute difference between $\tilde{M}^2_j$ and $M_j$ in each bin $j \in \left\{1, \ldots, (N - 1)/2\right\}$, which now yields
\begin{equation}
    \mathbb{E}_m\left[|\tilde{M}^2_j - M_j|\right] = \frac{1}{N} \left(
        j \frac{(N - j)}{N} + (N - j) \frac{j}{N} \right) = \frac{2 \, j \, (N - j)}{N^2},
\end{equation}
observing that there are $j$ outcomes with $|\tilde{M}^2_j - M_j| = (N - j) / N$ and $(N - j)$ outcomes with $|\tilde{M}^2_j - M_j| = j / N$ (see the bottom right grid in Fig.~\ref{fig:EMD_proof_sketch}). 
Exploiting the symmetry about the central bin again and averaging over the bins, we obtain
\begin{equation}
    \mathbb{E}_m\left[W_1(\tilde{m}^2, m)\right] = \frac{4}{N^3} \sum_{j=1}^{\frac{N-1}{2}} j \left(N - j\right) = \frac{N^2 - 1}{3 N^2}.
\end{equation}
Comparing the two strategies, we find that
\begin{equation}
    \mathbb{E}_m\left[W_1(\tilde{m}^2, m)\right] = \frac{4}{3} \  \mathbb{E}_m\left[W_1(\tilde{m}^1, m)\right],
\end{equation}
which confirms that predicting the median over all possible cumulative histogram realizations in each bin (Strategy 1) indeed leads to a smaller expected EMD as compared to the mean (Strategy 2) for $N > 1$.

\section{A Bimodal Toy Example}
\begin{figure}[!htb]
    \centering
    \resizebox{0.65\textwidth}{!}{
    \includegraphics{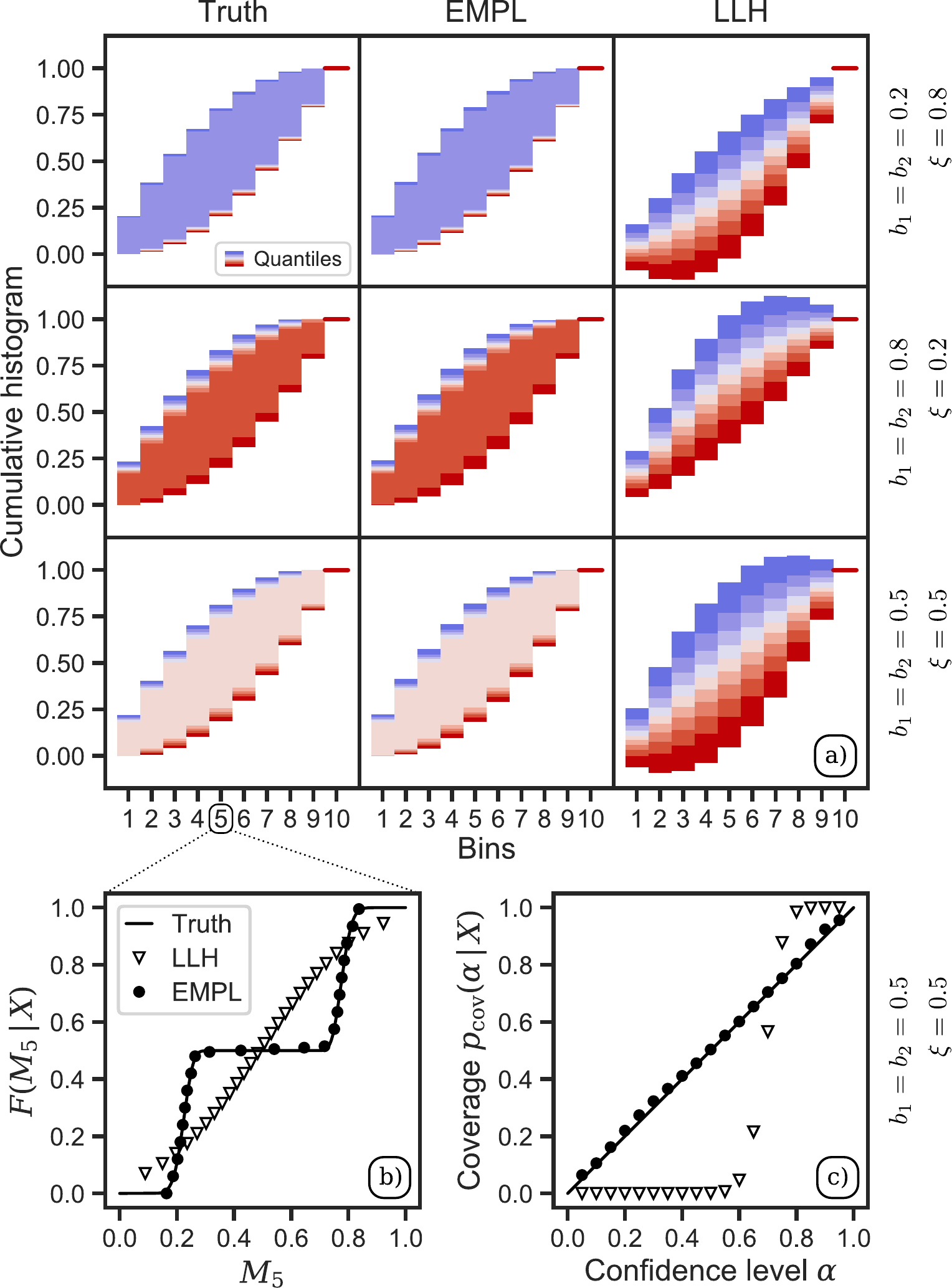}
    }
    \caption{Panel a): True and estimated distributions of the cumulative histograms, for three different inputs $X = (b_1, b_2, \xi)$ as specified next to each row. The colored regions show the $5 - 95\%$ quantiles in steps of $10\%$. The EMPL predictions (center) closely resemble the truth (left), whereas a NN trained to maximize the Gaussian log-likelihood within each bin (right) is clearly not able to capture the bimodal distribution. Panel b): CDF of the cumulative histogram value $M_j$ in bin $j = 5$ for the input $X = (0.5, 0.5, 0.5)$ considered in the bottom row of panel a). Panel c): Calibration plot for the same input $X = (0.5, 0.5, 0.5)$. The bin-averaged coverage probability $p_\text{cov}(\alpha \, | \, X)$ is given by the fraction of samples and bins that fall within the symmetric $\alpha$-interquantile range around the median. Perfectly calibrated uncertainties would lie on the identity line. 
    }
    \label{fig:bimodal}
\end{figure}
In the main body, we have presented the results of EMPL-trained NNs for three problems in different areas. In this section, we consider an additional toy example where the distribution of the cumulative histogram within each bin is \emph{bimodal}. 
The 3-dimensional NN input $X = (b_1, b_2, \xi)$ determines the width of each mode ($b_1, b_2 \in [0, 1]$) and the probability for the histogram label to follow the first mode ($\xi \in [0, 1]$; the probability to follow the second mode is thus $1 - \xi$). For the specific parameterization that we use to generate the histogram labels, as well as for the implementation details, we refer to our Github repository \href{https://github.com/FloList/EMPL}{\faGithub}. We use $N = 10$ histogram bins and train a MLP with 2 hidden layers consisting of 256 neurons each by minimizing the EMPL for 10,000 batch iterations at batch size 2,048. As in the football example, we compare our EMPL results with a NN trained by maximizing a Gaussian log-likelihood for the CDF in each bin.
\par Panel a) in Fig.~\ref{fig:bimodal} shows the true and estimated distributions of the cumulative histograms for three different NN inputs $X$, namely for narrow modes and a high probability for the first mode ($b_1 = b_2 = 0.2$, $\xi = 0.8$; top row), for wide modes and a high probability for the second mode ($b_1 = b_2 = 0.8$, $\xi = 0.2$; middle row), and for the symmetric case with moderately wide modes ($b_1 = b_2 = \xi = 0.5$; bottom row). The distribution predicted by the EMPL-trained NN closely resembles the truth for all considered inputs, whereas a Gaussian likelihood yields completely unsatisfactory results for these highly non-Gaussian uncertainties.
For the symmetric case in the bottom row ($X = (0.5, 0.5, 0.5)$), we also plot the CDF in bin~5, see panel b). The EMPL prediction agrees with the truth, while a Gaussian CDF clearly cannot account for the two modes of the distribution. Finally, we consider the calibration of the uncertainties by computing the bin-averaged coverage probability $p_\text{cov}(\alpha \, | \, X)$ as a function of the confidence level $\alpha$, conditional on the input $X = (0.5, 0.5, 0.5)$. We calculate the coverage probability as the fraction of samples and bins for which the true value falls within the $\alpha$-interquantile range symmetrically around the median. We use 65,536 histogram realizations for the evaluation, and we only average over those bins where the true cumulative histogram lies within $[\varepsilon, 1 - \varepsilon]$ for $\varepsilon = 10^{-5}$ in order to avoid biases due to numerical inaccuracies far below the magnitudes of interest.
We confirmed that the results are not sensitive to the exact choice of~$\varepsilon$. With the EMPL, the maximum deviation from perfect calibration ($p_\text{cov}(\alpha \, | \, X) = \alpha$) is $<$ 3\% for all the considered confidence levels $\alpha$, in contrast to $>$ 50\% with a bin-wise Gaussian log-likelihood loss function. This example demonstrates that the EMPL is able to accurately recover complex posterior distributions over plausible histograms, conditional on an input vector $X$.

\section{Tensorflow Implementation}
We provide a basic \texttt{Tensorflow} implementation of the EMPL below (tested with \texttt{Tensorflow 2.3} and \texttt{Python 3.8}), which can be adapted according to the specific use case. Note that the input to the function is given by the true and estimated \emph{density} histograms (not the cumulative histograms), which should be properly normalized, e.g. using a $\operatorname{softmax}$ activation function (however, the normalization is not checked by the function below).
\vspace*{-0.2cm}
\renewcommand{\theFancyVerbLine}{
  \sffamily\textcolor[rgb]{0.5,0.5,0.5}{\scriptsize\arabic{FancyVerbLine}}}
\begin{minted}[mathescape,
               numbersep=5pt,
               gobble=0,
               framesep=2mm,
               fontsize=\footnotesize,
               baselinestretch=1]{python}
import tensorflow as tf
def empl(m, m_tilde, tau=0.5, alpha=0.0, scope="empl"):
    """
        Computes the Earth Mover's Pinball Loss:
        :param m:          true histogram labels (shape: n_batch x n_bins)
        :param m_tilde:    histogram predictions (shape: n_batch x n_bins)
        :param tau:        quantile level tau in [0, 1]
        :param alpha:      smoothing parameter alpha (>= 0)
        :param scope:      scope name
        :returns           Earth Mover's Pinball Loss between the density histograms
                           m and m_tilde for the quantile level tau
    """
    assert len(m.shape) == len(m_tilde.shape) == 2, "Only 2D tensors are supported!"
    assert m.shape[0] == m_tilde.shape[0], "Batch dimensions do not agree!"
    assert m.shape[1] == m_tilde.shape[1], "Bin dimensions do not agree!"

    with tf.name_scope(scope):
        # Density histograms -> cumulative histograms
        M = tf.cumsum(m, axis=1)
        M_tilde = tf.cumsum(m_tilde, axis=1)
        
        # Compute difference
        delta = M_tilde - M

        # Non-smooth loss (default)
        if alpha == 0.0:
            mask = tf.cast(tf.greater_equal(delta, tf.zeros_like(delta)), delta.dtype) - tau
            loss = mask * delta

        # Smooth loss
        else:
            loss = -tau * delta + alpha * tf.math.softplus(delta / alpha)

        # Return mean over bins and batch
        return tf.reduce_mean(loss)
\end{minted}
\end{document}